\documentclass[runningheads]{llncs}
\usepackage{graphicx}
\usepackage{comment}
\usepackage{amsmath,amssymb} %
\usepackage{color}

\usepackage{cite}
\usepackage{url}
\usepackage{xspace}
\usepackage{hyperref}
\usepackage{booktabs}
\usepackage{multirow}
\usepackage[T1]{fontenc}
\usepackage{colortbl}

\usepackage{subfigure}

\usepackage{paralist}

\newcommand{\xhdr}[1]{\vspace{5pt}\noindent\textbf{#1}}

\newcolumntype{s}{>{\columncolor[gray]{.925}[.25\tabcolsep]}c}

\newcommand{\tabref}[1]{Table~\ref{#1}\xspace}
\newcommand{\figref}[1]{Figure~\ref{#1}\xspace}
\newcommand{\secref}[1]{Section~\ref{#1}\xspace}

\newcommand{\eg}{e.g.\xspace}
\newcommand{\ie}{i.e.\xspace}
\newcommand{\etal}{et~al.\xspace}
\newcommand{\vs}{vs.\xspace}

\newcommand{\myquote}[1]{{\it `#1'}}

\belowdisplayskip \abovedisplayskip

\newcommand{\csection}[1]{
    \vspace{-0.04in}
    \section{#1}
    \vspace{-0.03in}
}

\newcommand{\csubsection}[1]{
    \vspace{-0.04in}
    \subsection{#1}
    \vspace{-0.03in}
}

\begin{document}
\pagestyle{headings}
\mainmatter
\def\ECCVSubNumber{5672}  %

\title{Improving Vision-and-Language Navigation\\ with Image-Text Pairs from the Web} %

\author{
    Arjun Majumdar\inst{1} \and
    Ayush Shrivastava\inst{1} \and
    Stefan Lee\inst{3} \and \\
    Peter Anderson\inst{1}\thanks{Now at Google.} \and 
    Devi Parikh\inst{1,2} \and 
    Dhruv Batra\inst{1,2}
}

\authorrunning{A. Majumdar et al.}
\titlerunning{VLN-BERT}
\institute{
Georgia Institute of Technology \and
Facebook AI Research (FAIR) \and 
Oregon State University
}

\maketitle

\begin{abstract}
Following a navigation instruction such as \myquote{Walk down the stairs and stop at the brown sofa} requires embodied AI agents to ground scene elements referenced via language (\eg \myquote{stairs}) to visual content in the environment (pixels corresponding to \myquote{stairs}).

\vspace{1ex}
We ask the following question -- can we leverage abundant `disembodied' web-scraped vision-and-language corpora (\eg Conceptual Captions~\cite{sharma2018conceptual}) to learn visual groundings (what do \myquote{stairs} look like?) that improve performance on a relatively data-starved embodied perception task (Vision-and-Language Navigation)?
Specifically, we develop VLN-BERT, a visiolinguistic transformer-based model for scoring the compatibility between an instruction (\myquote{...stop at the brown sofa}) and a sequence of panoramic RGB images captured by the agent.
We demonstrate that pretraining VLN-BERT on image-text pairs from the web before fine-tuning on embodied path-instruction data significantly improves performance on VLN -- outperforming the prior state-of-the-art in the fully-observed setting by 4 absolute percentage points on success rate.
Ablations of our pretraining curriculum show each stage to be impactful -- with their combination resulting in further positive synergistic effects.

\end{abstract}

\csection{Introduction}
\label{sec:introduction}

\begin{figure}[t]
    \centering
    \includegraphics[width=.9\textwidth]{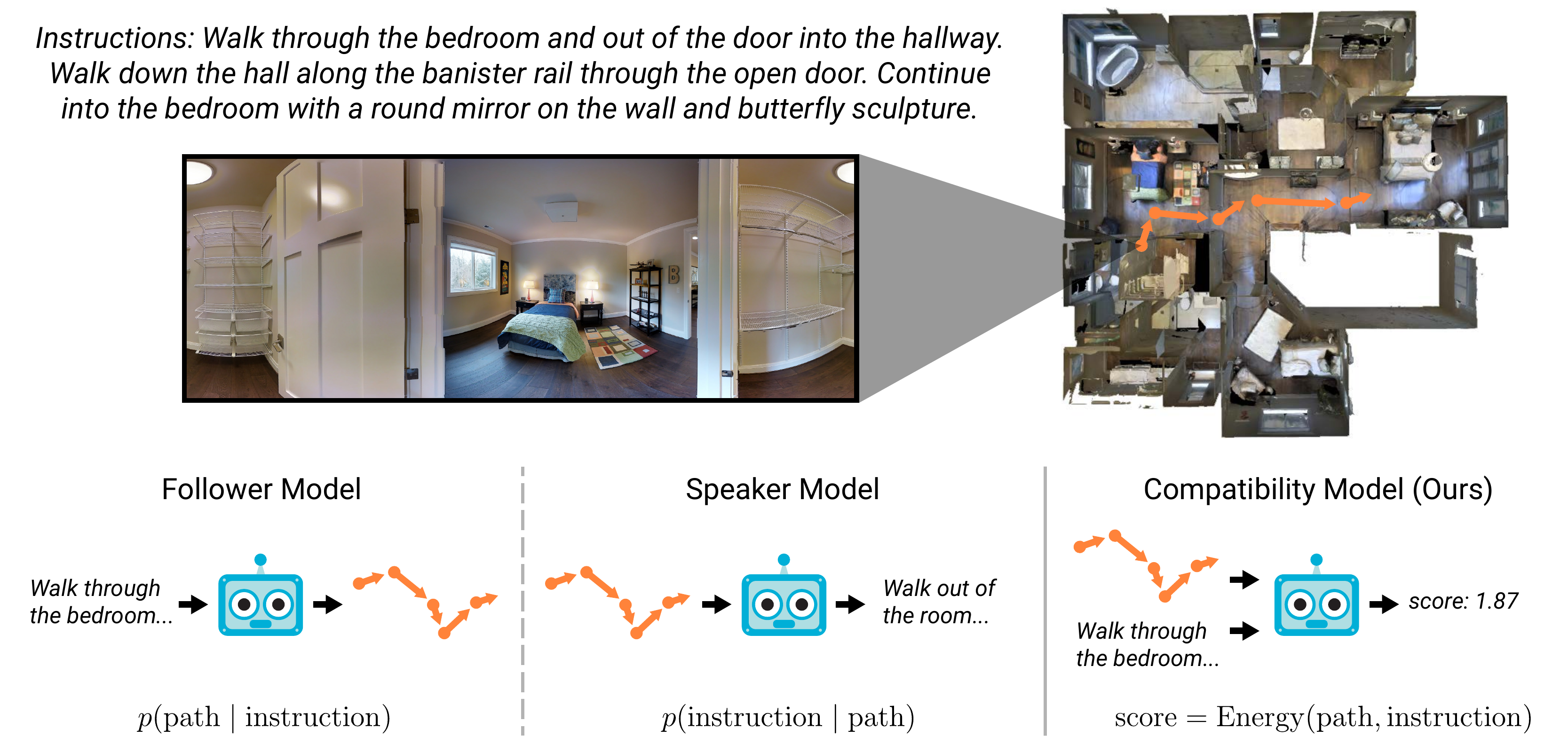}
    \caption{We propose a compatibility model (right) for path selection in vision-and-language navigation (VLN). In contrast to the follower (left) and speaker (center) models that have typically been used in prior work, our model takes a path and instruction pair as input and produces a score that reflects their alignment. Based on this model we describe a training curriculum that leverages internet data in the form of image-caption pairs to improve performance on VLN.}
    \label{fig:teaser}
\end{figure}

Consider the navigation instruction in Figure \ref{fig:teaser}, \myquote{Walk through the bedroom and out of the door into the hallway. Walk down the hall along the banister rail through the open door. Continue into the bedroom with a round mirror on the wall and butterfly sculpture.} ~In vision-and-language navigation (VLN)~\cite{anderson2018vision}, agents must interpret such instructions to navigate through photo-realistic environments. In this instance, the agent needs to exit the bedroom, walk past something called a \myquote{banister rail} and find the bedroom containing a \myquote{round mirror} and \myquote{butterfly sculpture.} But what if the agent has never seen a butterfly before, let alone a sculpture of one? To solve this task, an agent needs to determine if the visual evidence along a path matches the descriptions provided in the instructions. As such, the ability to ground references to objects and scene elements like \myquote{butterfly sculpture} and \myquote{banister rail} is central to success. Existing work has focused on learning this grounding solely from a task-specific training dataset of path-instruction pairs~\cite{wang2018look, fried2018, rcm, self-monitor, ma2019regretful, tactical-rewind, tan2019environment, anderson2019chasing} -- which are expensive, laborious, and time-consuming to collect at scale and thus tend to be relatively small (\eg the VLN dataset contains around 14k path-instruction pairs for training). As an alternative, we propose learning visual grounding from freely-available internet data, such as the web images with alt-text captured in the Conceptual Captions dataset~\cite{sharma2018conceptual}, containing around 3.3M image-text pairs.

\begin{figure}[t]
    \centering
    \includegraphics[width=1.0\textwidth]{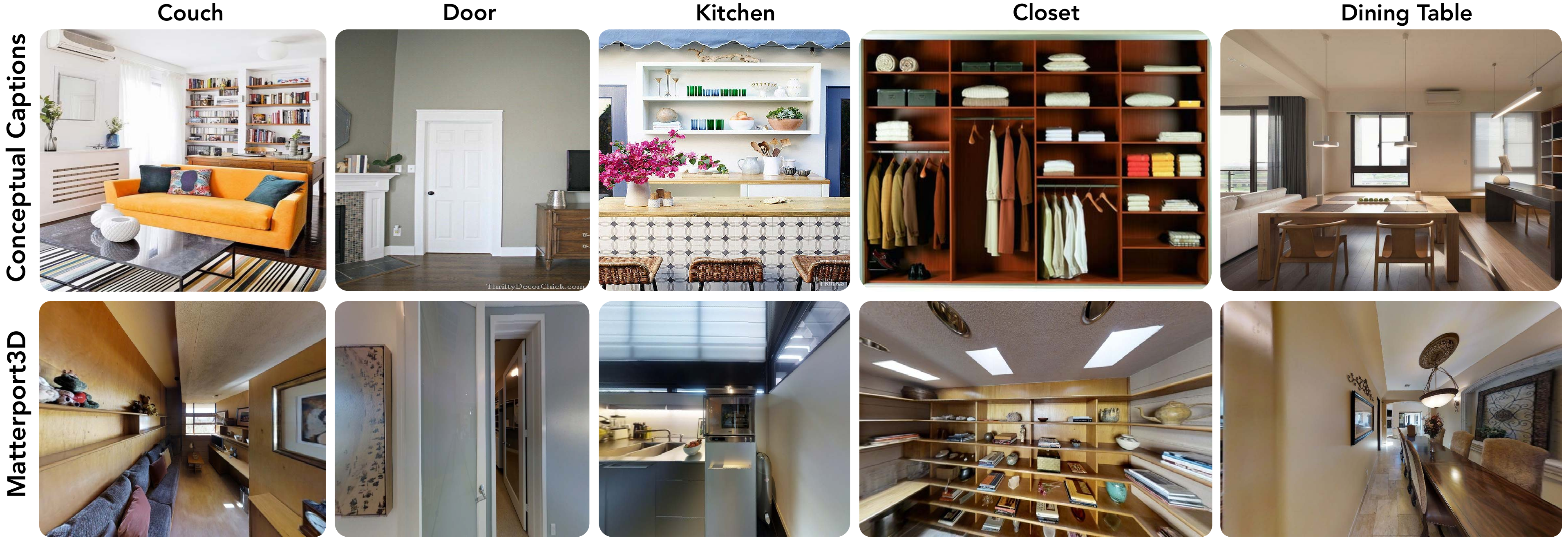}
    \caption{Images from the Conceptual Captions (CC)~\cite{sharma2018conceptual} (top) and Matterport3D (MP3D)~\cite{Matterport3D} (bottom) datasets illustrate the differences between the two domains. Images from CC are typically well-lit, well-composed and aesthetically pleasing, while for MP3D images (used in VLN) the framing depends on the position of the agent (\eg a couch (left) in CC is typically viewed head-on, whereas in MP3D they may be hidden to the side as an agent navigates past them).}
    \label{fig:domain-shift}
\end{figure}

Conceptually, transfer learning from large-scale web data to embodied AI tasks such as VLN is an attractive alternative to collecting more data. Empirically, however, the effectiveness of this strategy remains open to question -- would such a transfer even work? Unlike web images, which are highly-curated and stick closely to aesthetic biases, embodied data contains content and viewpoints that are not widely published online. For example, as shown in \figref{fig:domain-shift}, an embodied agent may perceive doors via a close-up view of a door frame rather than as a carefully composed image of a (typically closed) door. In VLN, image framing is a consequence of the agent's position rather than an aesthetic choice made by a photographer. Consequently, in this paper we investigate this question -- to what degree can webly-supervised visual grounding learned on static images be transferred to the embodied VLN task? Put more succinctly, can `disembodied' web data be used to improve visual grounding for embodied agents?

To answer this question, we introduce VLN-BERT, a joint visiolinguistic transformer-based compatibility model for scoring the alignment between an instruction and an agent's observations along a trajectory. We structure VLN-BERT to enable straight-forward transfer learning from a model from prior work on general visiolinguistic representation learning~\cite{lu2019vilbert}, and explore a training curriculum that incorporates both large-scale internet data and embodied path-instruction pairs. VLN-BERT is sequentially trained using 1) language-only data (Wikipedia and BooksCorpus~\cite{zhu2015aligning} as in BERT~\cite{devlin2018bert}), 2) web image-text pairs (Conceptual Captions~\cite{sharma2018conceptual} as in ViLBERT~\cite{lu2019vilbert}), and 3) path-instruction pairs from the VLN dataset~\cite{anderson2018vision}. Following this protocol the model progressively learns to represent language, then to ground visual concepts, and finally to ground visual concepts alongside action descriptions. We evaluate VLN-BERT on a path selection task in VLN, demonstrating that this training procedure leads to significant gains over prior work (4 absolute percentage points on leaderboard success rate).

\xhdr{Contributions.} Concretely, we make the following main contributions:\\[-0.75em]
\begin{compactitem}
    \item We develop VLN-BERT, a visiolinguistic transformer-based model for scoring path-instruction pairs. We show that VLN-BERT outperforms strong single-model baselines from prior work on the path selection task -- increasing success rate (SR) by 4.6 absolute percentage points.\\[-0.75em]
    \item We demonstrate that in an ensemble of diverse models VLN-BERT improves SR by 3.0 absolute percentage points on ``unseen'' validation, leading to a SR of 73\% on the VLN leaderboard (4 absolute percentage points higher than previously published work)\footnote{\href{https://evalai.cloudcv.org/web/challenges/challenge-page/97/leaderboard/270}{{\tt evalai.cloudcv.org/web/challenges/challenge-page/97/leaderboard/270}}}.\\[-0.75em]
    \item We ablate the proposed training curriculum, and find that each stage contributes significantly to the final outcome, with a cumulative benefit that is greater than the sum of the individual effects. Notably, we find that pretraining on image-text pairs from the web provides a significant boost in path selection performance -- improving SR by 9.2 absolute percentage points.\\[-0.75em]
    \item We provide qualitative evidence that our model learns to ground object references. Specifically, using gradient-based methods~\cite{simonyan2013deep} we visualize how image-region importance shifts under modifications to the instructions given to our model, demonstrating reasonable responses to these interventions. For example, if we modify the instruction \myquote{Walk down the stairs, then stop next to the fridge.} by removing \myquote{stop next to the fridge} we observe that image regions containing the fridge become less important.
\end{compactitem}

\csection{Related Work}
\label{sec:related}

\xhdr{Path Selection in VLN.}
In VLN~\cite{anderson2018vision}, an agent is required to follow a navigation instruction from a start location to a goal. While most existing works focus on the setting in which the test environments are previously unseen, many also consider the scenario in which the test environment is previously explored and stored in memory (\ie, fully observable). In this setting, a high-probability path is typically generated by performing beam search through the environment and ranking paths according to either: (1) their probability under a `follower' model~\cite{wang2018look, fried2018, rcm, self-monitor, ma2019regretful, tactical-rewind, anderson2019chasing}, as in Figure \ref{fig:teaser} (left), or (2) by how well they explain the instruction according to a `speaker' (instruction generation) model~\cite{fried2018,tan2019environment}, as in Figure \ref{fig:teaser} (center). In contrast, we use beam search with an existing agent model~\cite{tan2019environment} to generate a set of candidate paths, which we then evaluate using our discriminative path-instruction compatibility model, as in Figure \ref{fig:teaser} (right).  

\xhdr{Data Augmentation and Auxiliary Tasks in VLN.}
To compensate for the small size of existing VLN datasets, previous works have investigated various data augmentation strategies and auxiliary tasks. Many papers report results trained on augmented data including instructions synthesized by a speaker model~\cite{fried2018,self-monitor,ma2019regretful,tan2019environment}. Tan \etal~\cite{tan2019environment} use environmental dropout to mimic additional training environments to improve generalization. Li \etal~\cite{li2019robust} incorporate language-only pretraining using a BERT model. Several existing papers~\cite{rcm,huang2019multi} and one concurrent hitherto-unpublished work~\cite{hao2020towards} consider path-instruction compatibility as an auxiliary loss function or reward for VLN agents. We focus on path-instruction compatibility in the context of transfer learning from large-scale internet data, which has not been previously explored.

\xhdr{Vision-and-Language Pretraining.}
There has been significant recent progress towards learning transferable joint representations of images and text \cite{zhou2019unified,lu2019vilbert,tan2019lxmert,su2019vlbert,li2019unicoder,li2019visualbert}. Using BERT-like~\cite{devlin2018bert} self-supervised objectives and Transformer~\cite{vaswani2017attention} architectures, these models have achieved state-of-the-art results on multiple vision-and-language tasks by pretraining on aligned image-and-text data collected from the web~\cite{sharma2018conceptual} and transferring the base architecture to other tasks such as VQA~\cite{balanced_vqa_v2}, referring expressions~\cite{KazemzadehOrdonezMattenBergEMNLP14}, and caption-based image retrieval~\cite{chen2015microsoft}. However, these tasks are all based on single images. The extent to which these pretrained models can generalize from human-composed and curated internet images to embodied AI tasks has not been investigated. In this work we propose a training curriculum to handle potential domain-shift and augment a previous model architecture to process panoramic image sequences, extending the progress in vision-and-language to vision-and-language navigation (VLN).

\csection{Preliminaries: Self-Supervised Learning from the Web}
\label{sec:prelim}
Recent works have demonstrated that high-capacity models trained under self-supervised objectives on large-scale web data can learn strong, generalizable representations for both language and images \cite{devlin2018bert, zhou2019unified,lu2019vilbert,tan2019lxmert,su2019vlbert,li2019unicoder,li2019visualbert}. We build upon these works as a basis for transfer and describe them briefly here.

\xhdr{Language Modeling with BERT.}
The BERT \cite{devlin2018bert} model is a large transformer-based  \cite{vaswani2017attention} architecture for language modeling. The model takes as input sequences of tokenized words augmented with positional embeddings and outputs a representation for each. For example, a two sentence input could be written as
\begin{equation}
\mbox{\texttt{<CLS>~}} w^{(1)}_1, \dots, w^{(1)}_{L_1} \mbox{\texttt{~<SEP>~}} w^{(2)}_1, \dots, w^{(2)}_{L_2} \mbox{\texttt{~<SEP>~}}
\end{equation}
where \texttt{CLS}, and \texttt{SEP} are special tokens. To train this approach, \cite{devlin2018bert} introduce two self-supervised objectives -- \emph{masked language modelling} and \emph{next sentence prediction}. Given two input sentences from a text corpus, the masked language modelling objective masks out some percentage of tokens and tasks the model to predict their values given the remaining tokens as context. The next sentence prediction objective requires the model to predict whether the two sentences follow each other in the original corpus or not. BERT is then trained under these objectives on large language corpuses from the web (Wikipedia and BooksCorpus~\cite{zhu2015aligning}). This model forms the basis for both our approach and the visiolinguistic representation learning discussed next. 

{\sloppy
\xhdr{Visiolinguistic Representations Learning with ViLBERT.}
Extending BERT, ViLBERT~\cite{lu2019vilbert} (and a number of similar approaches \cite{zhou2019unified, lu2019vilbert, tan2019lxmert, su2019vlbert, li2019unicoder, li2019visualbert}) focuses on learning joint visiolinguistic representations from paired image-text data, specifically web images and their associated alt-text collected in the Conceptual Captions dataset \cite{sharma2018conceptual}. ViLBERT is composed of two BERT-like processing streams that operate on visual and textual inputs, respectively. The input to the visual stream is composed of image regions (generated by an object detector~\cite{Anderson2017up-down,renNIPS15fasterrcnn} pretrained on Visual Genome~\cite{krishnavisualgenome}) that act as ``words'' in the visual domain. Concretely, given a single image $I$ consisting of a set of image regions $\{v_1, \dots, v_k\}$ and a text sequence (\ie a caption) $w_1, \dots, w_L$, we can write the input to ViLBERT as the sequence
\begin{equation}
\mbox{\texttt{<IMG>~}} v_1, \dots, v_k \mbox{\texttt{~<CLS>~}} w_1, \dots, w_L \mbox{\texttt{~<SEP>~}}
\end{equation}
where \texttt{IMG}, \texttt{CLS}, and \texttt{SEP} are special tokens marking the different modality sub-sequences. The two streams are connected using co-attention~\cite{lu2016hierarchical} transformer layers, which attend from the visual stream over language stream and vice versa. Notably, the language stream of ViLBERT is designed to mirror BERT such that it can be initialized by a pretrained BERT model. After processing, the model produces a contextualized output representation for each input token.
}

In analogy to the training objectives in BERT, ViLBERT introduces the \emph{masked multimodal modelling} and \emph{multimodal alignment tasks}. In the first, a random subset of language tokens and image regions are masked and must be predicted given the remaining context. For image regions, this amounts to predicting a distribution over object classes present in the masked region. Masked text tokens are handled as in BERT.  The multimodal alignment objective trains the model to determine if an image-text pair matches, \ie if the text describes the image content. Individual token outputs are used to predict masked inputs in the masking objective, and the \texttt{IMG} and \texttt{CLS} tokens are used for the image-caption alignment objective. We build upon this model extensively in this work.

\csection{Approach}
\label{sec:approach}

We first describe the path selection setting in Vision-and-Language Navigation (\secref{sec:path}), then our proposed model architecture (\secref{sec:model}), and finally our transfer learning curriculum (\secref{sec:transfer}).

\begin{figure}[t]
    \centering
    \includegraphics[width=.9\textwidth]{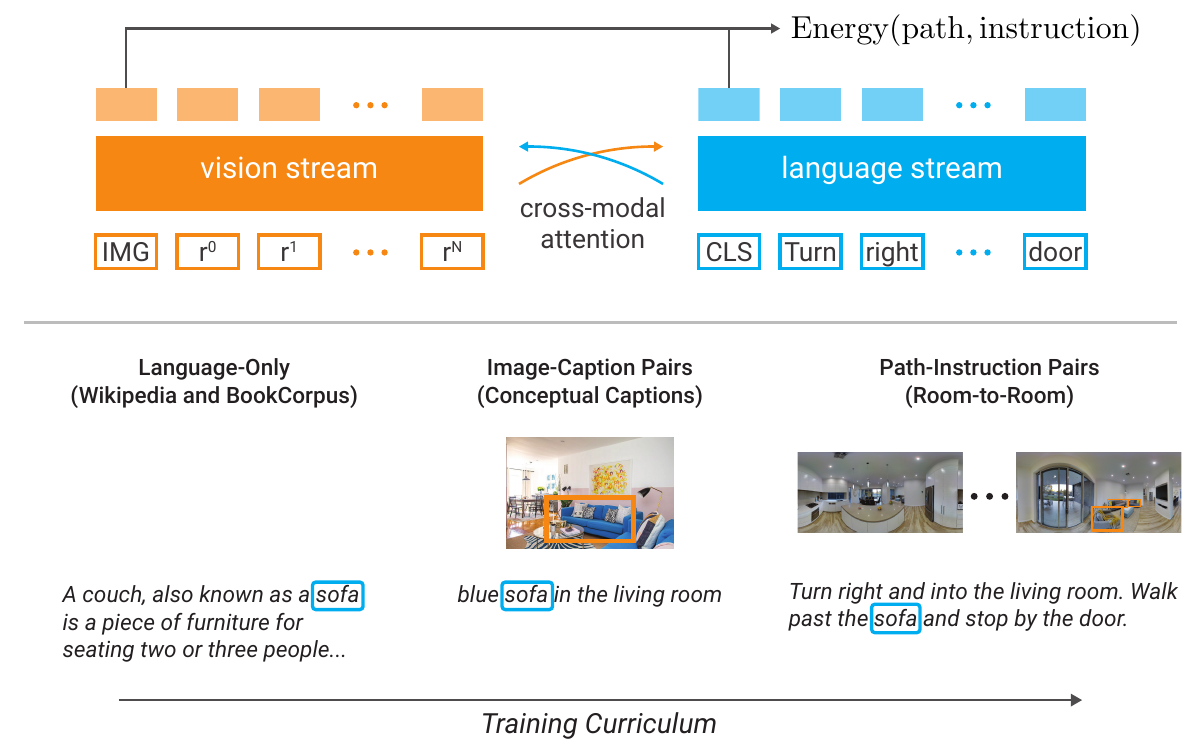}
    \caption{We propose VLN-BERT (top), a visiolinguistic transformer-based model similar to the model from~\cite{lu2019vilbert}, to process image regions from a sequence of panoramas and words from an instruction. We demonstrate that with the proposed training curriculum (bottom) visual grounding learned from image-text pairs from the web (center) can be transferred to significantly improve performance in VLN.}
    \label{fig:model}
\end{figure}

\csubsection{Vision-and-Language Navigation as Path Selection}
\label{sec:path}

In the Vision-and-Language Navigation (VLN) \cite{anderson2018vision} task, agents are placed in an environment specified as a navigation-graph $G = \{\mathcal{V}, \mathcal{E}\}$. Nodes $v \in \mathcal{V}$ represent different positions within the environment, and are represented by 360-degree panoramas taken at that viewpoint. Meanwhile, edges delineate navigable paths between panorama positions. The agent is provided with a navigation instruction $x$ that describes the shortest-path between a starting position $v_s$ and goal position $v_g$ (as illustrated in the bottom right of \figref{fig:model}). Agents are considered to have succeeded if they traverse a path $\tau = [v_s,v_1,v_2,\dots,v_N]$ with a final position $v_N$ that is within 3m of the goal  $v_g$.

Much of the work in VLN focuses on this problem as an exploration task in new environments; however, many practical deployments of robotic agents would be long-term in relatively fixed environments (\eg an assistant operating in a single home). In this paper, we consider the setting in which the environment is previously explored with the navigation-graph and panoramas are stored in the agent's memory. This setting has been studied in prior work \cite{fried2018,rcm,self-monitor,tactical-rewind,tan2019environment} and is operationalized by providing the agent unrestricted access to the Matterport3D Simulator \cite{anderson2018vision} during inference such that it can consider arbitrarily many valid paths originating from the starting position $v_s$, before selecting one to follow.

In this setting, the navigation task becomes one of identifying the path best aligned with the instructions. Concretely, given a set of valid paths $\mathcal{T}$ with the same starting position $v_s$ and an instruction $x$, the problem of navigation is to identify a trajectory $\tau^*$ such that 
\begin{equation}
    \tau^* = \operatornamewithlimits{argmax}_{\tau \in \mathcal{T}} ~~f(\tau, x)
\end{equation}
for some compatibility function $f$ that determines if the trajectory follows the instruction and terminates near the goal. The two major challenges are how to learn a compatibility function $f$ and how to efficiently search through the large set of possible paths. Given that our focus is on transfer learning, we address the first challenge within a simple path selection setting. Specifically, we consider a small set of paths $\mathcal{T^{\prime}}=\{\tau^1,\tau^2,\dots,\tau^M\}$ for each instruction, which are generated using beam-search with a greedy instruction-following agent \cite{tan2019environment}, and task $f$ with selecting the path that best aligns with the instruction from this set. Future work might explore how $f$ could be further used as a heuristic to efficiently search through the larger, exhaustive set of candidate paths $\mathcal{T}$.

\csubsection{Modeling Instruction-Path Compatibility}
\label{sec:model}

To formalize the task, we consider a function $f$ that maps a trajectory $\tau$ and an instruction $x$ to compatibility score $f(\cdot, \cdot)$. We model $f(\tau, x)$ as a visiolinguistic transformer-based model denoted as VLN-BERT. The architecture of VLN-BERT is structural similar to ViLBERT \cite{lu2019vilbert}; this is by design because it enables straight-forward transfer of visual grounding learned from large-scale web data. Specifically, we make a number of VLN-specific adaptations to ViLBERT, but they are all structured as augmentations (adding modules) rather than ablations (removing existing network components) so that pretrained weights can be transferred to initialize large portions of the model.

\xhdr{Representing Trajectories and Instructions.}
Predicting path-instruction compatibility requires jointly reasoning over a sequence of observations and a sequence of instruction words. As in prior work \cite{fried2018}, a trajectory is represented as a sequence of panoramic images (as in \figref{fig:model} bottom right) with positional information -- \ie $\tau = [ (I_1, p_1), \dots, (I_N, p_N) ]$ where $(I_i)$ are panoramas and $(p_i)$ are poses. Further, we represent each panorama $I_i$ as a set of image regions $\{r^{(i)}_1, \dots, r^{(i)}_K\}$. Let an instruction $x$ be a sequence of tokens $w_1, \dots, w_L$. We can thus write a path-instruction pair for VLN-BERT as the input sequence
\begin{equation}
\mbox{\texttt{<IMG>~}} r^{(1)}_1, \dots, r^{(1)}_K, \dots, \mbox{\texttt{<IMG>~}} r^{(N)}_1, \dots, r^{(N)}_K, \mbox{\texttt{~<CLS>~}} w_1, \dots, w_L \mbox{\texttt{~<SEP>~}}
\end{equation}
where \texttt{IMG}, \texttt{CLS}, and \texttt{SEP} are special tokens as before. 

The transformer models on which VLN-BERT (as well as BERT and ViLBERT) is based are inherently invariant to sequence order -- only representing interactions between inputs as a function of their values. The common practice to introduce this information is to add positional embeddings to the input token representations. For language, this is straight-forward and amounts to an index-in-sequence encoding. Panorama trajectories on the other have significantly more complex relationships. While the panoramas themselves are a sequence, there are also geometric relationships between them (\eg two panoramas being 1.2 meters apart at 10 degrees off north). Further, each individual image region not only has a position in the image (as modelled in ViLBERT)  but also an angle relative to the heading of an agent as it traverses the trajectory. These are important considerations for language-guided navigation -- after all, something on your left going one way is on your right if you go in the opposite direction. Being able to reason about the order of panoramas and the relative heading of image content is integral for following instructions like \myquote{Go down the hallway on the right then stop when you see a table on your left.}.

\begin{figure}[t]
    \centering
    \subfigure[\scriptsize Panoramic Spatial Information \label{fig:space}]{
    \includegraphics[height=0.68in]{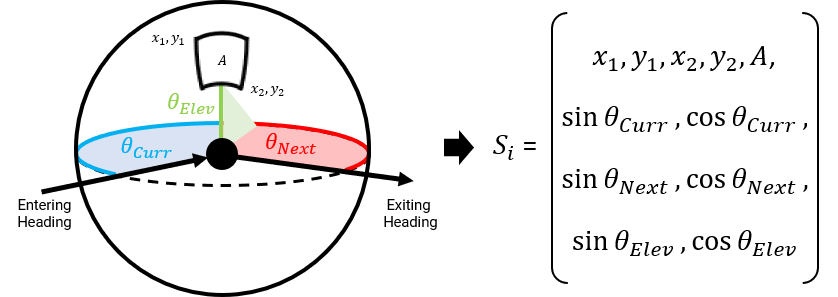}}\hfill%
    \subfigure[\scriptsize Overall Region Encoding \label{fig:sum}]{
    \includegraphics[height=0.68in]{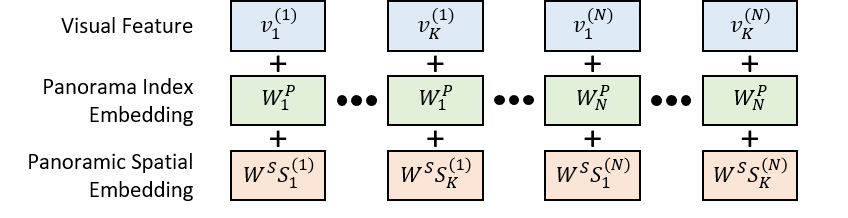}}
    \caption{We encode spatial information for each region to include not only the region position, but also its relation to the trajectory path (a). We form overall region encoding by summing visual features, an embedding indicating the index of the source panorama in the trajectory, and an embedding of panoramic spatial information (b).}
    \label{fig:encodings}
\end{figure}

To address this, as visualized in \figref{fig:space}, we encode the spatial location of each image region $r_i$ in terms of its location in the panorama (top-left and bottom-right corners in normalized coordinates as well as area of the image covered), its elevation relative to the horizon, and its heading relative to the agents current and next viewing directions.  All angles are encoded as $[\cos(\theta),\sin(\theta)]$. The resulting 11-dimensional vector $S_i$ is projected into 2048 dimensions using a learned projection $W^S$. To capture the sequential order of the panoramas within a trajectory, we project the scalar panorama index to 2048 dimensions using a learned embedding $W^P$. As shown in \figref{fig:sum}, the complete visual input representation for the image region is the element-wise sum of the visual features, panorama index embedding, and panoramic spatial embedding.

\xhdr{Extracting Image Regions from Panoramas.}
To extract the image regions $\{v^{(i)}_1, \dots, v^{(i)}_k\}$ for each panorama, we process the panoramas by first generating 600 $\times$ 600 pixel perspective projections using an 80 degree field of view at the 36 discrete heading and elevation directions used in previous work~\cite{anderson2018vision}. Similarly to ViLBERT, we then use the bottom-up attention model~\cite{Anderson2017up-down,renNIPS15fasterrcnn} pretrained on Visual Genome~\cite{krishnavisualgenome} to independently extract a set of image regions and features from each perspective image. Since the perspective images have substantial overlap we remove redundant regions within each panorama. First, we discard regions that are centred more than 20 degrees away from the center of the image (\ie we discard regions along the boarders). We assume that the discarded regions will be captured in a neighboring perspective image (spaced at 30 degree heading increments), with more visual context. Next, we examine pairs of image regions within each panorama in order of decreasing feature similarity. We discard the region in the pair with the lower bottom-up attention class detection score, until a maximum of 100 regions per panorama remain. We define similarity as the cosine distance between image features to which we add the absolute difference in region heading and elevation. Including heading and elevation differences ensures that visually similar features found in different regions of the panorama are unlikely to be classified as redundant. 

\xhdr{Training for Path Selection.}
To train VLN-BERT for path selection, we consider a 4-way multiple-choice task. Given an instruction $x$, we sample four trajectories out of which only one is successful $\{\tau^+_1, \tau^-_2, \tau^-_3, \tau^-_4\}$. We run VLN-BERT on each instruction-trajectory pair and extract their corresponding final representations. We denote these outputs for the \texttt{CLS} and the first \texttt{IMG} token as $h_{\mbox{\texttt{CLS}}}$ and $h_{\mbox{\texttt{IMG}}}$ respectively and compute a compatibility score $s_i$ as
\begin{equation}
    s_i = f(\tau_i,x) =  W \left(h^{(i)}_{\mbox{\texttt{CLS}}} \odot h^{(i)}_{\mbox{\texttt{IMG}}}\right)
\end{equation}
where $\odot$ denotes element-wise multiplication and $W$ is a learned transformation matrix. Scores, normalized via a softmax, are supervised with cross-entropy loss,
\begin{equation}
 \mathbf{p} = \mbox{softmax}(\mathbf{s})
\end{equation}
\begin{equation}
    \mathcal{L}\left(x,\{\tau^+_1, \tau^-_2,\tau^-_3,\tau^-_4\}\right) = \mbox{CrossEntropy}\left(\mathbf{p}, \mathbf{1}[\tau^+_1]\right)
\end{equation}
where $\mathbf{1}[\tau^+_1]$ is a 1-hot vector with mass at the index of $\tau^+$. At inference, we simply sort trajectories by their compatibility scores $s_i$.

\xhdr{Mining Negative Examples.}
We find that choosing an appropriate set of path-instruction pairs is critical to performance. Ideally, samples would span the space of all possible pairs, including hard negatives such as a hypothetical example where the agent must select between paths that end at two different \myquote{butterfly sculptures}. The question is how to find varied path-instruction pairs with semantically meaningful differences? We find that using beam search with an instruction-following model yields a diverse set of paths that are effective for training. Furthermore, the paths are conditioned on the instructions, and we find that in practice the incorrect paths often make semantically meaningful mistakes. Specifically, we sample up to 30 beams per instruction from the follower model of Tan \etal \cite{tan2019environment} and label the path as successful if it meets the VLN success criteria (\ie it ends $<3m$ from the goal). Finally, one positive and three negatives pairs are sampled uniformly at random for training.

\csubsection{Internet-to-Embodied Transfer Learning}
\label{sec:transfer}

While VLN-BERT can be trained from scratch, as described above we designed the model to specifically enable transfer learning from language \cite{devlin2018bert} and visiolinguistic \cite{lu2019vilbert} models trained on large-scale web corpora. This transfer is especially important in the VLN task which is relatively data-sparse (containing approximately 14k path-instruction pairs for training) and has a natural bias towards describing only objects present in training environments (\ie objects that are unique to the testing environments are never referenced in the training instructions). In this section, we describe a pretraining curriculum for transferring models learned on `disembodied' web data to the embodied VLN task.

We summarize the pretraining process in \figref{fig:model}. In total, we consider three stages focused on learning language, visual grounding, and action grounding.\\[-0.75em]
\begin{compactitem}[--]
\item \textbf{Stage 1: Language.} To capture strong language understanding capabilities, we initialize the language stream of our model with weights from a BERT~\cite{devlin2018bert} model trained on Wikipedia and the BooksCorpus~\cite{zhu2015aligning} under the masked language modelling and next sentence prediction objectives. Directly training on the path selection task after this stage is analogous to introducing a BERT encoder to represent instructions.\\[-0.75em]

\item \textbf{Stage 2: Visual Grounding.} Starting from a pretrained BERT model, Lu~\etal train both streams of ViLBERT on the Conceptual Captions dataset \cite{sharma2018conceptual} under the masked multimodal language modelling and multimodal alignment objectives. In this stage, we initialize model weights with a ViLBERT model trained in this manner. Transferring directly from this stage provides an initialization that can associate descriptions with image regions.\\[-0.75em]

\item \textbf{Stage 3: Action Grounding.} In the final stage, we pair paths and instructions from VLN and train the model under the masked multimodal modelling objective from \cite{lu2019vilbert}. While the previous stage learns to ground visual concepts, this stage additionally exposes the model to actions and their trajectory-based referents. For example, correctly predicting a masked instruction phrase like \myquote{turn \underline{\hspace{1cm}}} or \myquote{stop at the \underline{\hspace{1cm}}} requires the model to reason about the agent's path from the visual inputs and positional encodings.\\[-0.75em] 
\end{compactitem}

After these pretraining stages, we fine-tune our VLN-BERT model for path selection as described in the previous section.

\csection{Experiments}
\label{sec:experiments}

Our experiments primarily address following questions:\\[-0.75em]
\begin{compactenum}
    \item {\it Does pretraining on web image-text pairs improve VLN performance?}
    \item {\it How does the performance of VLN-BERT compare with strong baselines?}
    \item {\it Does VLN-BERT consider relevant image regions to produce alignment scores?}
\end{compactenum}

\csubsection{Dataset}
We conduct experiments using the Room-to-Room (R2R) navigation task~\cite{anderson2018vision} that was generated using the Matterport3D dataset~\cite{Matterport3D}. R2R contains human-annotated path-instruction pairs that are divided into training, seen and unseen validation, and unseen testing sets. To generate a dataset for path selection we run beam search on the instruction-follower model from \cite{tan2019environment}, to produce a set of up to 30 candidate paths for each instruction in R2R. We find that with a beam size of 30 over $99\%$ of the candidate sets contain one path that reaches the goal, which places an acceptable upper bound on path selection performance. In all of the experiments that follow, results are reported for selecting one path from the set of candidates.

\csubsection{Evaluation Metrics}
We compare the performance of different models using standard VLN metrics. Note that for path selection we calculate metrics using only the selected path, which corresponds with the pre-explored environment setting. However, for the VLN leaderboard results we follow the required approach of prepending the exploration path to the selected path (which affects path length based metrics).\\[-0.75em]
\begin{compactitem}
    \item {\bf Success rate (SR)} measures the percentage of selected paths that stop within 3m of the goal. In path selection this is our primary metric of interest.\\[-0.75em]
    \item {\bf Oracle Success rate (SR)} measures the percentage of selected paths with any position that passes within 3m of the goal.\\[-0.75em]
    \item {\bf Navigation error (NE)} measures the average distance of the shortest path from the last position in the selected path to the goal position.\\[-0.75em]
    \item {\bf Path length (PL)} measures the average length of the selected path -- a lower PL is preferred if the trajectory is successful.\\[-0.75em]
    \item {\bf Success rate weighted by path length (SPL)}, as defined in~\cite{anderson2018evaluation}, provides a measure of success normalized by the ratio between the length of the shortest path and the selected path.
\end{compactitem}

\csubsection{Training Baseline Models}

We compare with the follower and speaker models from~\cite{tan2019environment}, which achieve state-of-the-art performance on the VLN test set in an ensemble model setting. The only auxiliary dataset used to train these baseline models is ImageNet~\cite{russakovsky2015imagenet} (used to pretrain an image feature extractor). All of the other components are trained from scratch (including word embeddings). Data augmentation, via environmental dropout~\cite{tan2019environment}, is used to train the follower model and greatly improves performance. We report results using code and weights provided by Tan \etal~\cite{tan2019environment}.

\csubsection{Results}
\label{sec:results}

\setlength{\tabcolsep}{3pt}
\begin{table}[t]
\caption{We compare the contribution from the different stages of pretraining. We find that stage 2 and 3 both contribute significantly to task performance (improving Val Unseen Success Rate (SR) by $\sim$15-20 absolute percentage points over the no pretraining baseline -- rows 3 and 4 \vs 1), with their full combination providing further synergistic gains (row 5). Notably, skipping the visual grounding stage (but still doing the others) results in a 9 absolute percentage point drop in Val Unseen SR (compare rows 4 and 5) -- demonstrating the importance of internet-to-embodied transfer of visual grounding.}
\vspace{-1.5em}
\label{tab:ablation}
\begin{center}
\resizebox{\columnwidth}{!}{
\begin{tabular}{lr ccc ccccs ccccs}
\toprule
&& \multicolumn{3}{c}{\bf Pretraining Stage} & \multicolumn{5}{c}{\bf Val Seen} & \multicolumn{5}{c}{\bf Val Unseen} \\
\cmidrule(l{4pt}r{4pt}){3-5}
\cmidrule(l{4pt}r{4pt}){6-10}
\cmidrule(l{4pt}r{0pt}){11-15}
& \texttt{\#} & \scriptsize\shortstack{ Language\\Only} & \scriptsize\shortstack{ Visual\\Grounding} & \scriptsize\shortstack{ Action\\Grounding} & PL & NE $\downarrow$ & SPL $\uparrow$ & OSR $\uparrow$ & SR $\uparrow$ & PL & NE $\downarrow$ & SPL $\uparrow$ & OSR $\uparrow$ & SR $\uparrow$\\
\midrule
\multirow{6}{*}{VLN-BERT} & 1 & \multicolumn{3}{c}{(no pretraining)} & 10.78 & 6.78 & 0.35 & 54.22 & 37.55 & 10.29 & 6.81 & 0.27 & 50.62 & 30.52 \\
\cmidrule(l{0pt}r{0pt}){3-15}
& 2 & \checkmark & & & 10.33 & 4.89 & 0.55 & 69.31 & 58.73 & 9.59 & 5.47 & 0.41 & 57.34 & 45.17 \\
& 3 & \checkmark & \checkmark & & 10.42 & 4.48 & 0.58 & 71.57 & 62.16 & 9.70 & 4.96 & 0.45 & 62.79 & 49.64 \\
& 4 & \checkmark & & \checkmark & 10.51 & 4.28 & 0.60 & 72.65 & 63.82 & 9.81 & 5.05 & 0.46 & 62.75 & 50.02 \\
\cmidrule(l{0pt}r{0pt}){3-15}
& 5 & \checkmark & \checkmark & \checkmark & 10.28 & 3.73 & 0.66 & 76.47 & 70.20 & 9.60 & {\bf 4.10} & {\bf 0.55} & {\bf 69.22} & {\bf 59.26} \\
\bottomrule
\end{tabular}
}
\end{center}
\end{table}
\setlength{\tabcolsep}{1.4pt}

\xhdr{Does pretraining on web image-text pairs improve VLN performance?}
To answer this question we dissect our proposed training curriculum as indicated in \tabref{tab:ablation}, and find that in general each stage of training does contribute to performance. First, we find that our model has limited performance learning from scratch, achieving only 30.5\% SR (compared with the 54.7\% SR achieved by the speaker model from~\cite{tan2019environment}). However, language-only pretraining, which corresponds to initializing our model with BERT~\cite{devlin2018bert} weights, improves performance substantially to 45.2\% SR (an improvement of 14.7 absolute percentage points) -- indicating that language understanding plays an important role in VLN.

Next, we find that both pretraining on image-text pairs from the Conceptual Captions~\cite{sharma2018conceptual} (visual grounding) and pretraining on path-instruction pairs from VLN~\cite{anderson2018vision} (action grounding) similarly improve success rate (by 4.5 and 4.9 absolute percentage points, respectively) when used independently. However, when the two pretraining stages are combined in series the improvement jumps to 14.1 absolute percentage points in success rate or 9.2 absolute percentage points over the next best setting. The substantial level of improvement that results from our full training curriculum suggests that not only does pretraining on webly-supervised image-text pairs from~\cite{sharma2018conceptual} improve path selection performance, but it also constructively supports the action grounding stage (Stage 3) of pretraining.

\setlength{\tabcolsep}{3pt}
\begin{table}[ht]
\caption{Results comparing VLN-BERT with the follower and speaker from~\cite{tan2019environment}. Notably, in the ensemble models setting, combining VLN-BERT with the speaker and follower results in a 3 absolute percentage point improvement in Val Unseen Success Rate (SR) over the next best three-model ensemble (compare rows 6 and 7).}
\vspace{-1.5em}
\label{tab:main-result}
\begin{center}
\scriptsize
\resizebox{\columnwidth}{!}{
    \begin{tabular}{lrl ccccs c ccccs}
        \toprule
        &&& \multicolumn{5}{c}{\bf Val Seen} & & \multicolumn{5}{c}{\bf Val Unseen} \\
        \cmidrule(l{2pt}r{4pt}){4-8}
        \cmidrule(l{4pt}r{0pt}){9-14}
         & \texttt{\#} & Re-ranking Model & PL & NE $\downarrow$ & SPL $\uparrow$ & OSR $\uparrow$ & SR $\uparrow$ & & PL & NE $\downarrow$ & SPL $\uparrow$ & OSR $\uparrow$ & SR $\uparrow$\\
        \midrule
        \multirow{3}{*}[-0.25em]{\rotatebox{90}{\shortstack{\scriptsize Single\\\scriptsize Models}}}
        & 1 &\hspace{.1em} follower~\cite{tan2019environment} & 10.40 & 3.68 & 0.62 & 74.12 & 65.10 & & 9.57 & 5.20 & 0.49 & 58.79 & 52.36 \\
        & 2 &\hspace{.1em} speaker~\cite{tan2019environment} & 11.19 & 3.80 & 0.56 & 77.25 & 60.69 & & 10.71 & 4.25 & 0.49 & {\bf 72.07} & 54.66 \\
        \cmidrule(l{4pt}r{0pt}){3-14}
        & 3 &\hspace{.1em} VLN-BERT & 10.28 & 3.73 & 0.66 & 76.47 & 70.20 & & 9.60 & {\bf 4.10} & {\bf 0.55} & 69.22 & {\bf 59.26} \\
        \midrule
        \multirow{3}{*}{\rotatebox{90}{\scriptsize\shortstack{\scriptsize Ensemble\\\scriptsize Models}}}
        & 4 &\hspace{.1em} speaker + follower~\cite{tan2019environment} & 10.69 & 2.72 & 0.70 & 82.94 & 74.22 & & 10.10 & 3.32 & 0.63 & 76.63 & 67.90 \\
        & 5 &\hspace{.1em} speaker + follower + follower & 10.73 & 2.72 & 0.71 & 83.33 & 74.71 & & 10.12 & 3.22 & 0.64 & 77.56 & 69.14 \\
        & 6 &\hspace{.1em} speaker + follower + speaker & 10.77 & 2.45 & 0.73 & 85.98 & 76.86 & & 10.17 & 2.99 & 0.65 & 79.28 & 70.58 \\
        \cmidrule(l{4pt}r{0pt}){3-14} 
        & 7 & \hspace{.1em} speaker + follower + VLN-BERT & 10.61 & 2.35 & 0.78 & 86.57 & 81.86 & & 10.00 & {\bf 2.76} & {\bf 0.68} & {\bf 81.91} & {\bf 73.61} \\
        \bottomrule
    \end{tabular}
}
\end{center}
\end{table}
\setlength{\tabcolsep}{1.4pt}

\setlength{\tabcolsep}{3pt}
\begin{table}[ht]
\begin{center}
\caption{Leaderboard results on Test Unseen for methods using beam search.}
\label{tab:leaderboard}
\scriptsize
\resizebox{0.75\columnwidth}{!}{
\begin{tabular}{l ccccs}
\toprule
& \multicolumn{5}{c}{\bf Test Unseen} \\
\cmidrule(l{4pt}r{0pt}){2-6}
Re-ranking Model & PL & NE $\downarrow$ & SPL $\uparrow$ & OSR $\uparrow$ & SR $\uparrow$ \\
\midrule
Speaker-Follower \cite{fried2018} & 1,257 & 4.87 & 0.01 & 96 & 53 \\
Tactical Rewind \cite{tactical-rewind} & 197 & 4.29 & {\bf 0.03} & 90 & 61 \\
Self-Monitoring \cite{self-monitor} & 373 & 4.48 & 0.02 & 97 & 61 \\
Reinforced Cross-Modal Matching \cite{rcm} & 358 & 4.03 & 0.02 & 96 & 63 \\	
Environmental Dropout \cite{tan2019environment} & 687 & 3.26 & 0.01 & {\bf 99} & 69 \\
Auxiliary Tasks$\dagger$ \cite{zhu2019auxilary} & 41 & 3.24 & 0.21 & 81 & 71 \\
\midrule
VLN-BERT & 687 & {\bf 3.09} & 0.01 & {\bf 99} & {\bf 73} \\
\bottomrule
\vspace{-.5em}\\
\multicolumn{6}{c}{$\dagger$indicates unpublished/concurrent work}
\end{tabular}
}
\end{center}
\end{table}
\setlength{\tabcolsep}{1.4pt}

\xhdr{How does VLN-BERT compare with strong baseline methods?}
The results in \tabref{tab:main-result} compare path selection performance of VLN-BERT with the state-of-the-art speaker and follower models from~\cite{tan2019environment}. We evaluate path selection using the set of up to 30 candidate paths generated with beam search using the follower from~\cite{tan2019environment}. For the follower model results this amounts to taking the top beam from the candidate set. In the single model setting we see that VLN-BERT, trained with our full curriculum, achieves 59.3\% SR, which is 4.6 absolute percentage points better than either of the other two methods.

In the pre-explored setting, the speaker and follower models are typically combined in an ensemble to further improve path selection performance~\cite{fried2018,tan2019environment}. The two models are typically combined as a linear combination using a hyperparameter $\alpha$ that is selected through grid search on the val unseen split of R2R~\cite{fried2018}. In the ensemble models section of \tabref{tab:main-result}, the speaker + follower line (row 4) represents our execution of the state-of-the-art ensemble model from~\cite{tan2019environment}. In rows 5-7, we consider three model ensembles composed of a speaker, follower, and one additional model combined using two hyperparameters $\alpha$ and $\beta$ (again selected through grid search on val unseen). We find that adding another (randomly seeded) speaker or follower model yields modest improvements of 1.2 and 2.7 absolute percentage points in SR (rows 5 and 6). In contrast, adding VLN-BERT results in a 5.7 absolute percentage point boost in SR (row 7), which is 3.0 absolute percentage points higher on success rate than the next best ensemble.

In \tabref{tab:leaderboard} we report results on the VLN test set via the VLN leaderboard. In the leaderboard setting we use a three-model ensemble that includes a speaker, follower, and VLN-BERT. The ensemble achieves a success rate of 73\%, which is 4 absolute percentage points greater than previously published work~\cite{tan2019environment}, and 2 absolute percentage points greater than concurrent, unpublished work~\cite{zhu2019auxilary}.

\begin{figure}[ht]
    \centering
    \includegraphics[width=1.0\textwidth]{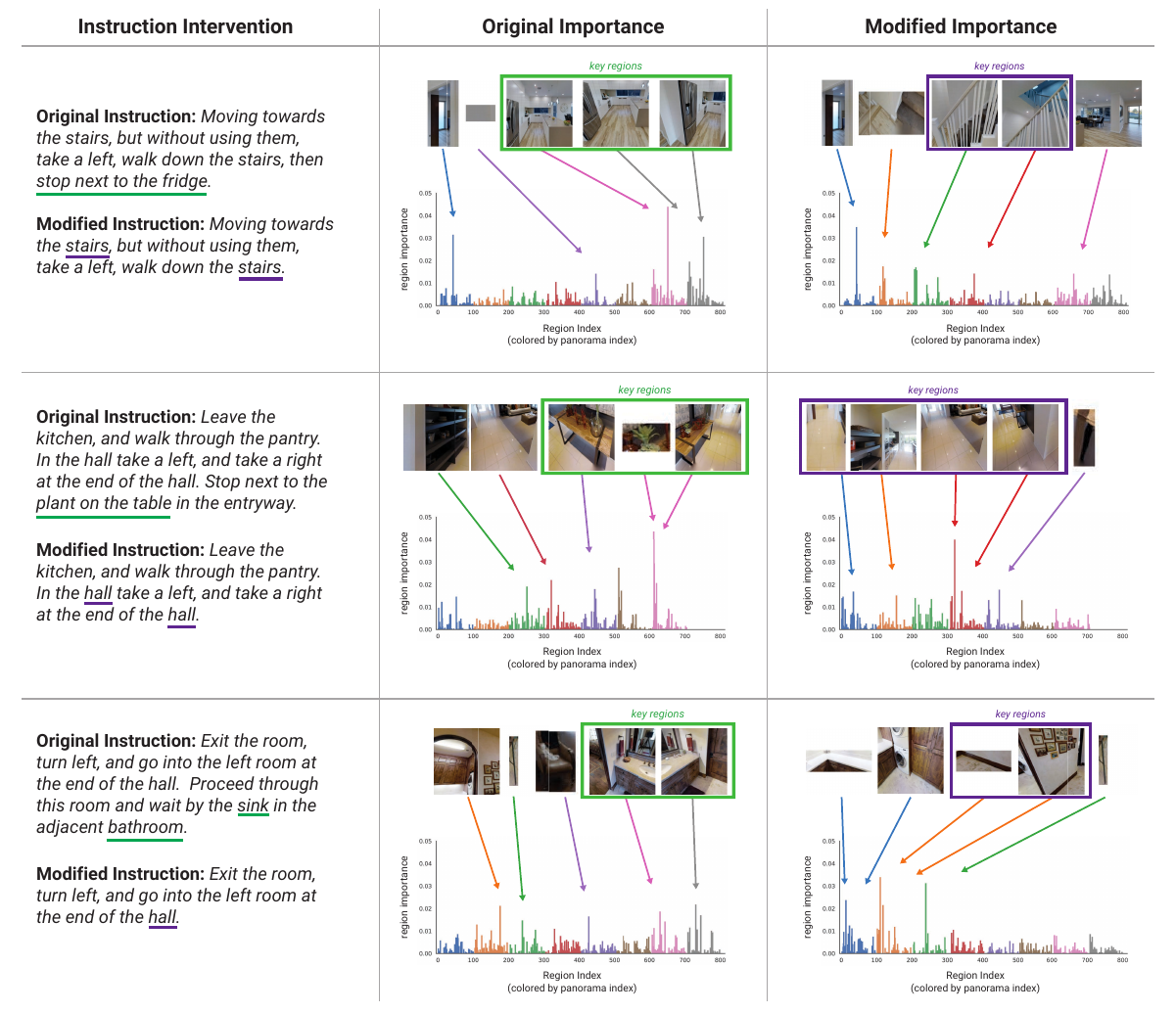}
    \caption{We compare region importance histograms for three path-instruction pairs under perturbations of the instruction -- removing a phrase or sentence. The region importance histogram is calculated by taking the gradient of the compatibility score with respect to the image region features. The images above each histogram correspond to the most influential regions (\ie the peaks in the histogram). The underlined instruction phrases correspond with the regions outlined in green and purple to provide a qualitative assessment of the visiolinguistic grounding learning by VLN-BERT. In the first example, removing the reference to the \myquote{fridge} (in green) shifts the importance to other regions along the path (\ie the \myquote{stairs} in purple), suggesting that VLN-BERT considers visually relevant image regions to score path-instruction pairs.}
    \label{fig:analysis}
\end{figure}

\xhdr{Does VLN-BERT consider relevant image regions to produce alignment scores?}
\label{sec:analysis}
One motivation behind our proposed approach is to improve the grounding of object references in VLN instruction. To test whether this actually happens and to gain insight into the improved performance of VLN-BERT, we visualize which parts of the visual input affect the compatibility score. We perform this analysis using a simple gradient-based visualization technique~\cite{simonyan2013deep}. We take the gradient of our learned score $f(x,\tau)$ with respect to the feature representation for each region from each panorama, and sum this 2048-dimensional gradient vector over the feature dimension to produce a scalar measure of region importance. To gain deeper insight we analyze how the region importance changes when the instruction is perturbed by removing parts of the description.

Three examples of this analysis are illustrated in \figref{fig:analysis}. The left panel provides the original and modified versions of the instruction with high-importance regions highlighted in green and purple, respectively. In each row, the original instruction is modified by removing a phrase that references a high-importance region (e.g. in the first row \myquote{then stop next to the fridge} is removed). In the middle and right panels, the region importance histograms and top 5 regions illustrate which parts of the visual input most influence the compatibility score. For example, in the first row regions containing a \myquote{fridge} are important for the original instruction, whereas for the modified instruction the importance shifts to the \myquote{stairs}. Independently, the center and right panels demonstrate that VLN-BERT produces compatibility scores based on relevant image regions. Furthermore, by comparing the top 5 regions before and after the instruction is modified, we see that the grounding of object references appropriately shifts under linguistic interventions. This analysis provides qualitative evidence that VLN-BERT properly learns to associate instruction phrases with image regions.

\csection{Conclusion}
\label{sec:discussion}

In this work, we demonstrated internet-to-embodied transfer of visual concept grounding -- leveraging large-scale image-text data from the web to improve a discriminative path-instruction alignment model for VLN. In our path re-ranking setting, this model improves over prior work and our ablations show each stage of our transfer curriculum contributes significantly.
\section*{Acknowledgements}

The Georgia Tech effort was supported in part by NSF, AFRL, DARPA, ONR YIPs, ARO PECASE, Amazon. The views and conclusions contained herein are those of the authors and should not be interpreted as necessarily representing the official policies or endorsements, either expressed or implied, of the U.S. Government, or any sponsor.

\clearpage
\bibliographystyle{splncs04}
\bibliography{references}

\clearpage
\appendix
\csection{Supplementary Material}

In this section we provide further implementation details for VLN-BERT (\secref{sec:implementation}) and additional qualitative analysis of performance (\secref{sec:qualitative-examples}) and the pretraining curriculum (\secref{sec:qualitative-pretraining}).
Code will be provided to reproduce the results of all experiments.

\csubsection{Implementation Details}
\label{sec:implementation}

The experiments described in \secref{sec:experiments} utilize a 12-layer $\text{BERT}_{\text{BASE}}$~\cite{devlin2018bert} architecture for both the vision and language streams in the model. Following~\cite{lu2019vilbert}, we use 6 cross-modal attention layers to connect the two streams. To operationalize the language-only pretraining stage (Stage 1), VLN-BERT is initialized with $\text{BERT}$ weights that result from pretraining on English Wikipedia and BooksCorpus~\cite{zhu2015aligning}. Similarly, for the visual grounding stage (Stage 2), VLN-BERT is initialized with $\text{ViLBERT}$ weights, which result from pretraining on the Conceptual Captions~\cite{sharma2018conceptual} dataset. For action grounding pretraining and path-selection finetuning (Stage 3) we use path-instruction pairs from the Room-to-Room~\cite{anderson2018vision} dataset. During this stage, models are trained with the Adam optimizer with a learning rate of 4e-5 and a batch size of 64. We use a learning rate schedule with a linear warmup and cooldown. We train for 50 epochs in pretraining and 20 epochs in finetuning. During finetuning, we utilize early stopping based on the success rate on the unseen split of the validation set. Using 8 Titan X GPUs Stage 3 of training takes approximately 66 hours.

\csubsection{Qualitative Examples of Success and Failure}
\label{sec:qualitative-examples}

This section provides qualitative examples of the path selection performance of VLN-BERT using the full training curriculum described in \secref{sec:transfer}. To gain insight at the region-level, we estimate region importance using the gradient-based visualization technique described in \secref{sec:analysis} (\ie importance is calculated as the sum of the gradient of the model's output score with respect to the features for each region). In all examples, the model selects one path from a set of up to 30 candidate paths for a given set of instructions. The selected path is successful if it terminates within 3m of the goal location.

\begin{figure}[t]
    \centering
    \includegraphics[width=.9\textwidth]{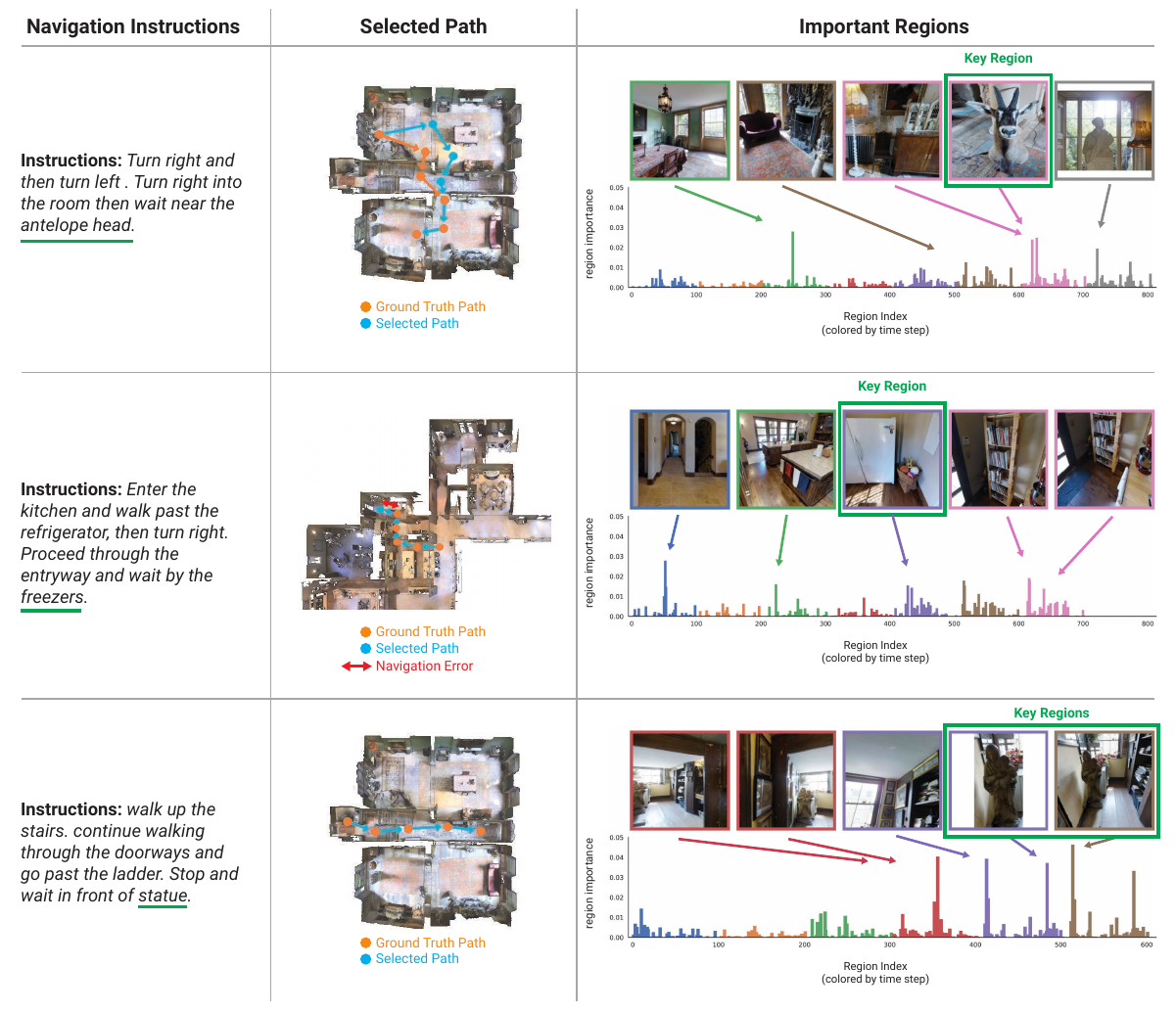}
    \caption{Examples of successful paths selected by VLN-BERT (middle -- blue), with ground truth paths (middle -- orange) and navigation errors (middle -- red) provided for comparison. The right column illustrates the top 5 regions that influence the model's predictions -- importance is determined by taking the gradient of the score for a path-instruction pair with respect to the input region features (as in \secref{sec:analysis}). A qualitative assessment of accurate visiolinguistic grounding (in green) highlights phrases that rarely occur in VLN training dataset: \myquote{antelope head} (0~occurrences), \myquote{antelope} (3~occurrences), \myquote{freezer(s)} (0~occurrences), \myquote{statue} (105~occurrences). These results suggest that VLN-BERT has effectively learned to transfer grounding from image-text pairs from the web to the embodied task of VLN.}
    \label{fig:positive}
\end{figure}

Three examples of successful path selection are illustrated in \figref{fig:positive}. In the first example, VLN-BERT selects a path that does not initially follow the ground truth path, but correctly stops at the goal location. In this example, the model accurately grounds the phrase \myquote{antelope head}, which does not appear in the VLN~\cite{anderson2018vision} training dataset (the term \myquote{antelope} appears 3 times). In the second and third examples, the selected paths closely match the ground truth, and the top 5 regions include key objects mentioned in the instructions -- \myquote{freezers} and \myquote{statue}. The term \myquote{freezers} (with and without the \myquote{s}) does not occur the VLN training dataset, and \myquote{statue} appears 105 times. These examples suggest that VLN-BERT is able to transfer visual grounding learned on the image-text pairs in the Conceptual Captions~\cite{sharma2018conceptual} dataset to the embodied task of path selection.

\begin{figure}[t]
    \centering
    \includegraphics[width=.9\textwidth]{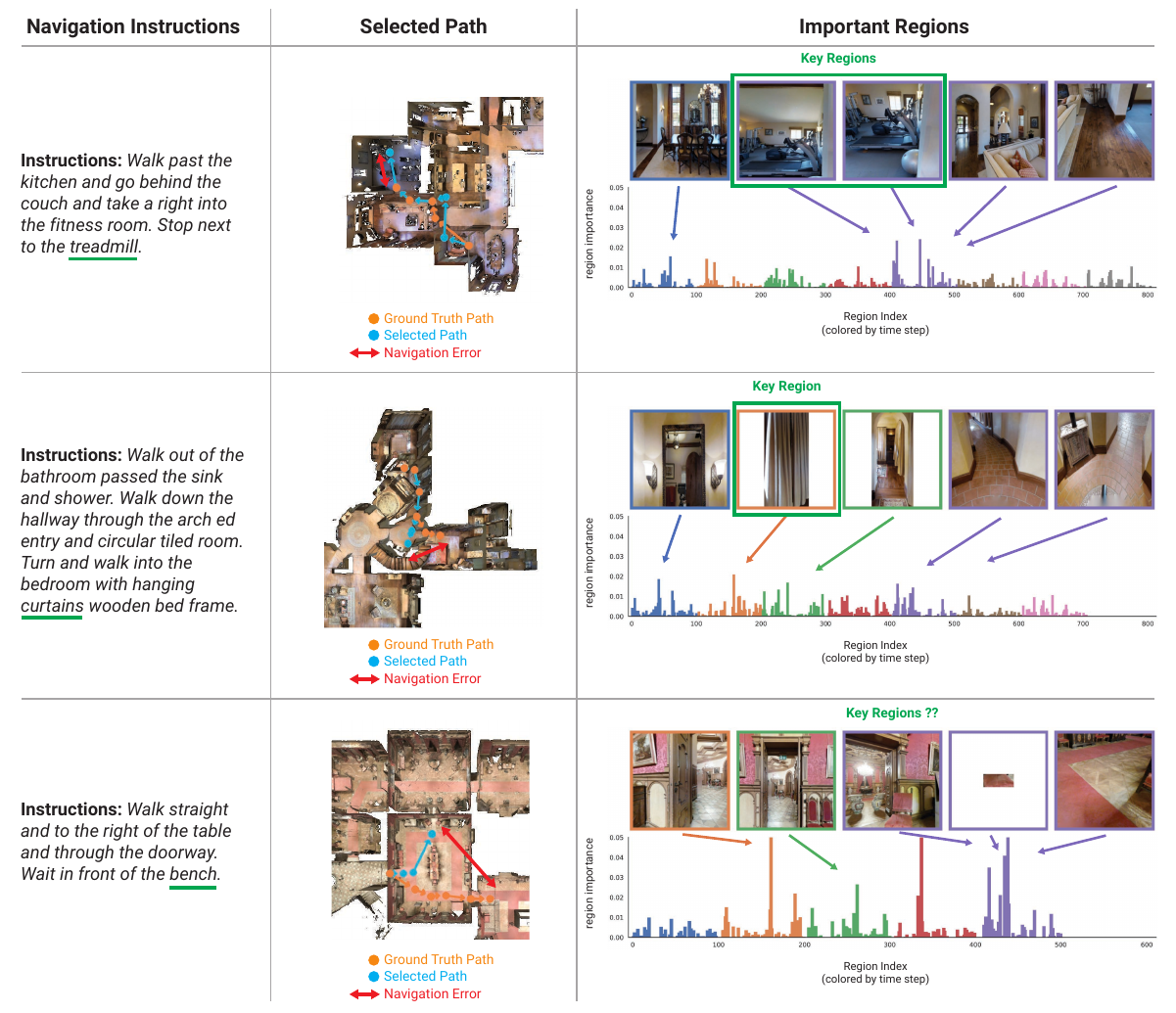}
    \caption{Examples of unsuccessful paths selected by VLN-BERT (middle -- blue), with ground truth paths (middle -- orange) and navigation errors (middle -- red) provided for comparison. The right column illustrates the top 5 regions that influence the model's predictions -- importance is determined by taking the gradient of the score for a path-instruction pair with respect to the input region features (as in~\secref{sec:analysis}). In the first row, the selected path goes past the goal location, but the visual grounding appears accurate. In the second row, the selected path does not enter the correct bedroom and the visual grounding appears to identify the wrong curtains. In the final example the model fails completely -- selecting a path going to the left (not right) of the table.}
    \label{fig:negative}
\end{figure}

Three unsuccessful examples are shown in \figref{fig:negative}. In each example the characteristics of the errors are qualitatively different. In the first row, VLN-BERT selects a path that does not stop at the correct goal location. However, the top 5 regions include key visual landmarks mentioned in the instruction (\eg \myquote{treadmill} and \myquote{sofa}). In contrast, in the second row, VLN-BERT selects a path that does not reach the goal bedroom. In this instance, the top 5 regions include \myquote{curtains} from a different location, which may have led to this particular error. In the last row every aspect of the selected path seems mismatched from the instructions: the path goes to the left of the table not right and objects mentioned in the instructions are missing from the top 5 regions.

\csubsection{Qualitative Analysis of the Pretraining Curriculum}
\label{sec:qualitative-pretraining}

\begin{figure}[t]
    \centering
    \includegraphics[width=.9\textwidth]{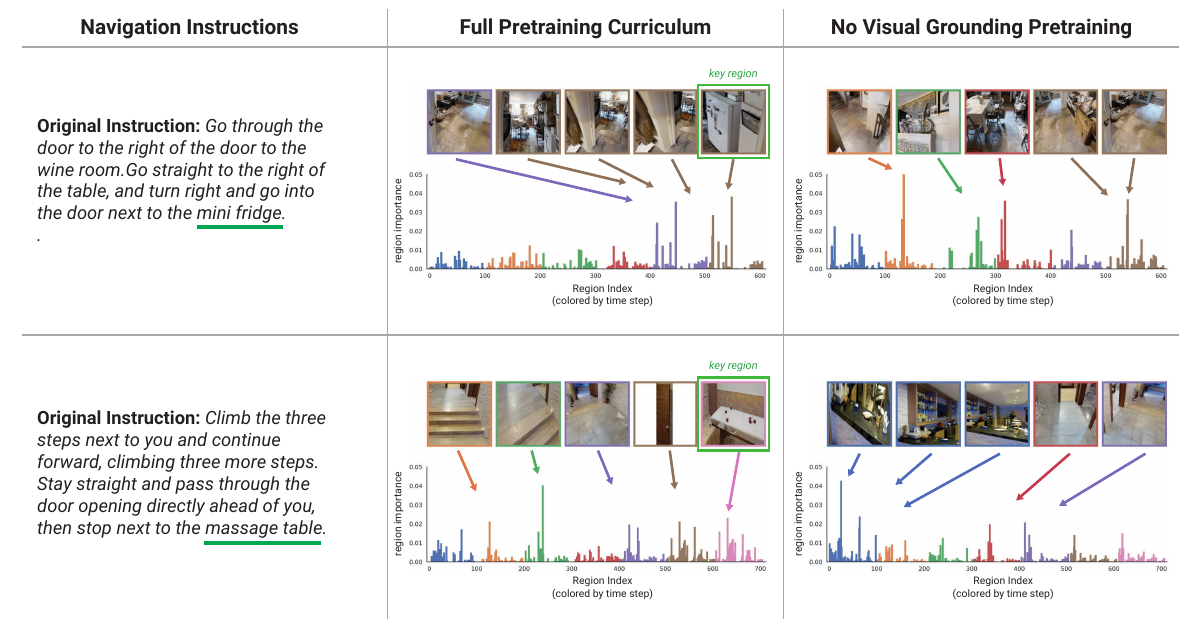}
    \caption{Comparison of pretraining with the proposed curriculum (middle) vs. omitting the visual grounding stage (right). Region importance histograms are estimated as in~\secref{sec:analysis}. The top 5 regions correspond with the successful path selected by VLN-BERT when pretrained with the full curriculum. Without the visual grounding pretraining stage (right), the model fails to ground phrases that rarely occur in the VLN training dataset (\eg \myquote{mini fridge} (1 occurrence) and \myquote{massage table} (29 occurrences)). Furthermore, in both cases without the visual grounding pretraining stage, VLN-BERT selects an unsuccessful path (not illustrated) for the given instructions.}
    \label{fig:pretrain-compare}
\end{figure}

In section~\secref{sec:results} we demonstrated that the visual grounding pretraining stage (Stage 2) quantitatively improves performance. In~\figref{fig:pretrain-compare} we qualitatively compare the visual grounding that is learned with and without stage 2 of pretraining. In the first example, the model trained without stage 2 of pretraining (right) fails to ground the phrase \myquote{mini fridge}, which only occurs 1 time in the VLN training dataset. Similarly, in the second example, without stage 2 pretraining, the model fails to ground the phrase \myquote{massage table} (29~occurrences in the VLN training dataset). In both cases, the model without stage 2 pretraining selects an unsuccessful path. In contrast, when trained with the full curriculum, VLN-BERT correctly grounds these key phrases and selects successful paths for these two examples.

\end{document}